\documentclass{article}

\usepackage[preprint]{neurips_2023}

\usepackage[utf8]{inputenc} 
\usepackage[T1]{fontenc}    
\usepackage{hyperref}       
\usepackage{url}            
\usepackage{booktabs}       
\usepackage{amsfonts}       
\usepackage{nicefrac}       
\usepackage{microtype}      
\usepackage{xcolor}         

\usepackage{wrapfig}
\usepackage{amsmath}
\usepackage{adjustbox}
\usepackage{todonotes}
\usepackage{booktabs}

\title{DDxT: Deep Generative Transformer Models for Differential Diagnosis}

\author{
    Mohammad Mahmudul Alam \\
    University of Maryland, Baltimore County \\
    \texttt{m256@umbc.edu} \\
        \And
    Edward Raff \\
    Booz Allen Hamilton \\
    Laboratory for Physical Sciences \\
    University of Maryland, Baltimore County \\
    \texttt{Raff\_Edward@bah.com} \\
        \And 
    Tim Oates \\
    University of Maryland, Baltimore County \\
    \texttt{oates@umbc.edu} \\
        \And 
    Cynthia Matuszek \\ 
    University of Maryland, Baltimore County \\
    \texttt{cmat@umbc.edu}
}

\begin{document}

\maketitle

\begin{abstract}
Differential Diagnosis (DDx) is the process of identifying the most likely medical condition among the possible pathologies through the process of elimination based on evidence. An automated process that narrows a large set of pathologies down to the most likely pathologies will be of great importance. The primary prior works have relied on the Reinforcement Learning (RL) paradigm under the intuition that it aligns better with how physicians perform DDx. In this paper, we show that a generative approach trained with simpler supervised and self-supervised learning signals can achieve superior results on the current benchmark. The proposed Transformer-based generative network, named DDxT, autoregressively produces a set of possible pathologies, i.e., DDx, and predicts the actual pathology using a neural network. Experiments are performed using the DDXPlus dataset. In the case of DDx, the proposed network has achieved a mean accuracy of $99.82\%$ and a mean F1 score of $0.9472$. Additionally, mean accuracy reaches $99.98\%$ with a mean F1 score of $0.9949$ while predicting ground truth pathology. The proposed DDxT outperformed the previous RL-based approaches by a big margin. Overall, the automated Transformer-based DDx generative model has the potential to become a useful tool for a physician in times of urgency.
\end{abstract}

\section{Introduction}
Differential Diagnosis (DDx) is referred to the process of systematically identifying a disease from a possible set of pathologies through the process of elimination based on a patient's medical history and physical examinations \cite{cook2020higher}. During a clinical process, a doctor asks several questions about the patient’s symptoms and antecedents (medical history). Based on the response, possible differential diagnoses are narrowed down. If there is uncertainty about the underlying condition, then a medical examination is performed or additional tests are suggested. Given a patient’s information and symptoms, an automated system that narrows down the possible pathologies if not identifying the exact one will be of great benefit. In particular, such improvements could help lower-performing doctors or those in under-resourced communities obtain better diagnostic outcomes \cite{tafti2022technology}. Moreover, in times of emergency, an automated system that has access to the patient’s medical history and current conditions will be quite valuable.
\par 
In recent years, automated diagnosis systems using machine learning have increasingly developed \cite{zhang2019automatic, marques2020automated, faris2021intelligent, irmak2022covid}. Existing works have demonstrated the potential of such automated systems in performing complete blood count (CBC) test~\cite{alam2019machine}, syndrome detection \cite{luo2021knowledge}, coronavirus, heart disease, and diabetes detection \cite{kumar2021efficient}, and more. Previous work such as \emph{Diaformer} \cite{chen2022diaformer} also demonstrated success in automated diagnosis using a sequence of explicit and implicit symptoms of a disease. But what is lacking is the details of the symptoms, the patient's previous medical history, and relevant information such as age, and gender. In a DDx process, a doctor would consider all of these information. 
\par 
In this paper, an automated DDx system is proposed using Transformer \cite{vaswani2017attention}, named \emph{DDxT}, that would take a sequence of all the patient’s necessary information as input to perform DDx by autoregressively generating a set of most likely pathologies and finally, predict the ground truth pathology using a neural network. This sequence of patient information will contain age, gender, medical history, and evidence, i.e., symptoms. Transformer architecture is employed since it is currently state-of-the-art for sequence generation \cite{brown2020language}. Asking questions to a patient and acquiring information can easily be done through an automated system. The challenging part is to make an intelligent decision based on the acquired information which will be addressed in this paper. This will be beneficial not only during the time of emergency but also as an assistive tool to the doctor during the diagnosis process.

\section{Related Works} \label{sec:related}
Recent works have demonstrated the feasibility of the machine learning-based automated diagnosis system. Such work is presented by \citep{huang2022transformer} where a Transformer-based model is utilized for the differential diagnosis. To perform the task, multi-modal magnetic resonance imaging (MRI) is utilized where a sequence of the brain and spinal cord MRI is processed by the Transformer. Their model performed considerably better than the previous approaches, however, the work is limited only to the diagnosis of demyelinating diseases. 
Likewise, \cite{sunena2021automated} presented an ensemble approach for automated diagnosis. Their approach involves multiple deep learning-based approaches where the final prediction is the ensemble of all the predictions. On the other hand, \cite{alam2019machine} proposed an automated complete blood count (CBC) test system which is a very common test in medical diagnosis. Their approach employed YOLO \cite{redmon2016you} object detection algorithm for blood cell detection. 
\par 
An image-based classifier is a powerful tool for automated diagnosis. Such a system is presented by \cite{papandrianos2022deep} where convolutional neural networks (CNN) are utilized to diagnose Coronary Artery Disease (CAD) using Myocardial Perfusion Imaging (MPI). Their system utilizes and compares performance on pre-trained VGG-16 \cite{simonyan2014very} and DenseNet-121 \cite{huang2017densely} architectures. In the same fashion, \cite{lv2020deep} developed an automated classification system for fungal keratitis. Their system uses ResNet \cite{he2016deep} architecture with fungal hyphae images for binary classification of fungal keratitis. 
Similarly, \cite{marques2020automated, marques2022ensemble} both employed EfficientNet \cite{tan2019efficientnet} for the detection of COVID-19 using X-ray images and Malaria classification from the blood smear images, respectively. In a slightly different manner, \cite{philippi2023vision} adopts vision transformer-based Swin-UNETR \cite{hatamizadeh2022swin} model to automatic retinal lesion segmentation from spectral-domain optical coherence tomography (SD-OCT) images.
\par 
The rest of the paper is organized as follows. \autoref{sec:approach} will cover the proposed method including a description of the dataset, network, and training procedure. Next, \autoref{sec:eval} will highlight the results and compare the proposed method to the RL agent-based methods. Finally, we conclude in \autoref{sec:conclusion} with a discussion of the limitations of our approach.

\section{Proposed Method} \label{sec:approach}
In this paper, differential diagnosis will be performed using a generative Transformer which will take a sequence of patient information as input and predict a sequence of most likely pathologies as differential diagnosis, and finally, the most likely pathology will be predicted using a classifier. In the following subsections, a brief description of the dataset, proposed network architecture, and training process will be discussed.

\subsection{Dataset} 
For differential diagnosis, along with evidence, i.e., symptoms, patient's antecedents (medical history) and personal details such as age and sex are necessary information. DDXPlus \cite{tchango2022ddxplus} dataset is such a dataset that contains synthetically generated 1.3M patient information where each sample contains patient details, evidence, ground truth differential diagnoses, and the ground truth condition. The dataset has a total of 49 pathologies that cover various age groups, sexes, and patients with a broad spectrum of medical history. We note that our work assumes the fidelity of the data since obtaining diagnostic data and medical history from patients comes at high expense, legal hurdles, ethics review, and slow collection rate ~\cite{cancerPain}. Such challenges are beyond the scope of our study. 
\par 
The dataset is preprocessed so that it can be processed by the Transformer. Each patient's information in the dataset contains age, sex, initial evidence, evidence (symptoms), ground truth differential diagnosis, and ground truth pathology. The age is categorized into 8 groups in the following way: \texttt{[less than 1)}, \texttt{[1-4]}, \texttt{[5-14]}, \texttt{[15-29]}, \texttt{[30-44]}, \texttt{[45-59]}, \texttt{[60-74]}, and \texttt{[above 75]}. Sex is represented by \texttt{M} for male and \texttt{F} for female. The Initial and rest of the evidence were acquired by back-and-forth questioning with a patient. Differential diagnosis contains a set of likely pathologies with a probability score for each pathology based on the evidence. Therefore, the ground truth DDx output sequence is organized in descending order of the probability score of each pathology, i.e., the order of prediction is significant and the pathology with a higher probability needs to be predicted first. Finally, the ground truth pathology is what the patient actually has.
\par 
Special tokens are incorporated to facilitate the learning process. Particularly, \texttt{<bos>} indicates the beginning of the sequence, \texttt{<sep>} token is used to separate each type of information, and to indicate the end of sequence \texttt{<eos>} is used. Since all sequences need to be equal in size, \texttt{<pad>} token is used in shorter sequences to fill out the sequence up to the maximum length, and longer sequences are truncated. Each patient's information is preprocessed as follows. First \texttt{<bos>} is used to initiate a sequence. Next, age, sex, initial evidence, and evidence all are stacked together using \texttt{<sep>} in between. Finally, the end of the sequence is indicated by \texttt{<eos>} token. To cover the unknown words in special circumstances, \texttt{<unk>} token is included in the vocabulary.

\subsection{Network Architecture}

\begin{wrapfigure}[19]{l}{0.5\textwidth}
\vspace{-14pt}
\centerline{\includegraphics[width=0.5\textwidth]{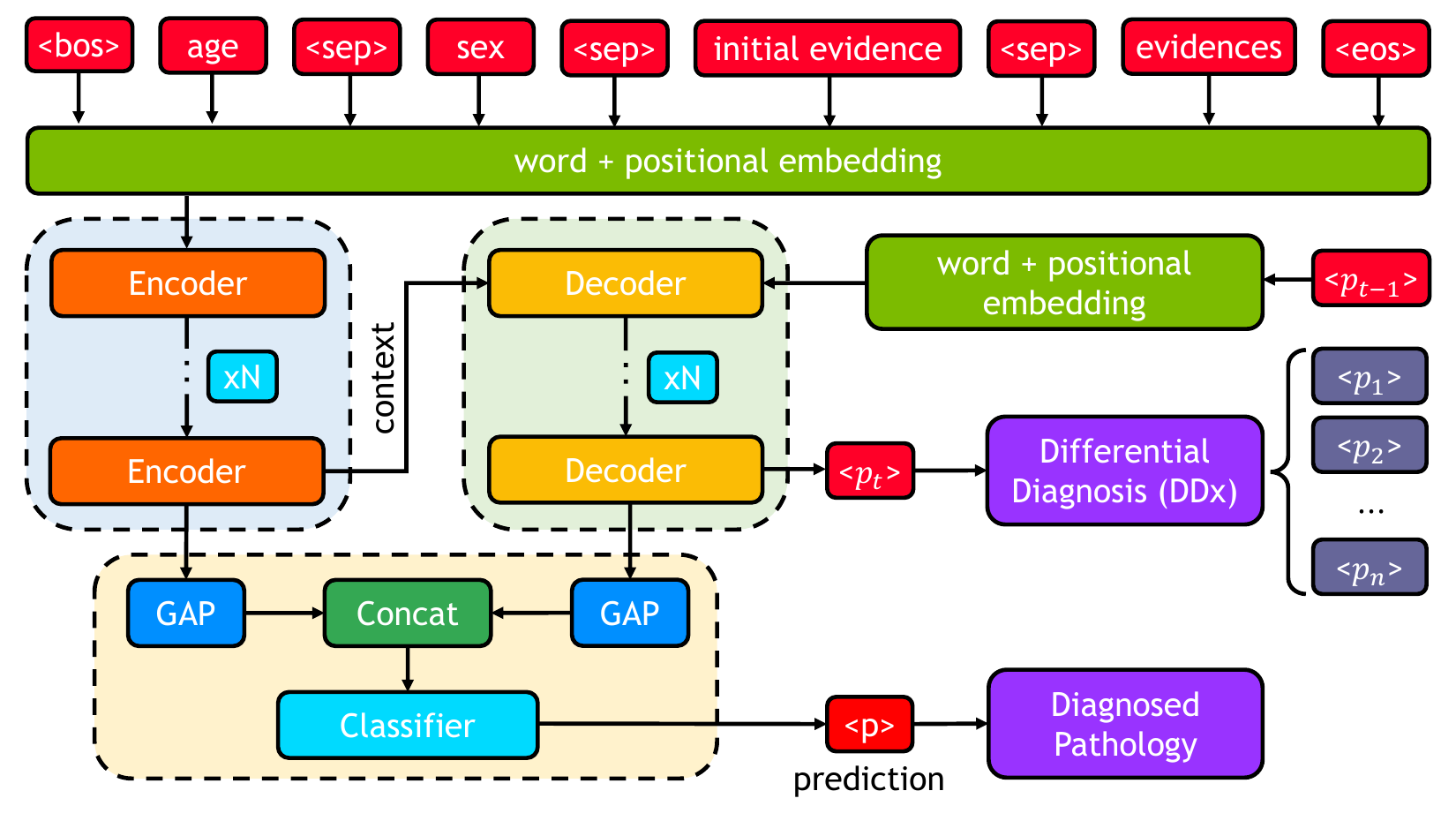}}
\caption{The block diagram of the proposed deep generative network architecture. The encoder blocks are shadowed with blue and the decoder blocks are shadowed with green. The classifier section is shadowed with orange. DDx is the set of $n$ pathologies from <$p_1$> to <$p_n$> and the classifier predicts the diagnosed pathology <$p$>.} 
\label{fig:block_diagram}
\end{wrapfigure}

After preprocessing the dataset, a vocabulary is built using all the unique tokens. The input string is split into words and using the generated vocabulary, each word is replaced with the associated index of the word in the vocabulary. The encoder vocabulary length is 436 and the decoder vocabulary length is 54 (49 pathologies + 5 special tokens). Next, these integer values are utilized to gather the associated word embedding and added with positional embedding so that the order of the word in a sequence is recognized by the network. Similarly, the decoder input tokens are also preprocessed, and word and positional embedding are applied.  The Transformer architecture consists of encoder and decoder blocks. Each of the blocks contains a self-attention mechanism, a brief description of which is provided in \autoref{a:appendix}. The encoder will process the patient’s information and feed the context to the decoder. The decoder will be initialized with the $p_0$=\texttt{<bos>} token which will iteratively take previously generated pathology $p_{t-1}$ as input and use tokens $p_0$ to $p_{t-1}$ to generate a new possible pathology $p_t$ until it reaches \texttt{<eos>} token. Both encoder and decoder are repeated $N (N = 6)$ times which will help recognize richer context. The decoder output is the DDx, a sequence of most likely pathologies. 
\par 
The final layer of the encoder holds the processed context information of the evidence, i.e., symptoms and relevant patients’ information, and the final layer of decoder holds the information of all the possible likely pathologies. Therefore, combining both features will be quite advantageous in predicting the actual pathology. As a result, Global average pooling (GAP) is applied to both the encoder and decoder features, concatenated, and fed to a classifier. The classifier is a two-layer neural network. The first layer contains the same number of features as the encoder or decoder, and the second layer has the same number of logits as the number of classes in the dataset. Both layers are preceded by layer normalization \cite{ba2016layer}. In between the layers, GELU activation \cite{hendrycks2016gaussian} is used. The classifier predicts the ground truth pathology among the most likely DDx. The full block diagram of the network architecture is presented in Figure \ref{fig:block_diagram}.

\subsection{Training} 
During training, the input size of the encoder and decoder must be fixed. Therefore, the maximum sequence length for the encoder is set to be $80$, and the maximum sequence length for the decoder is set to be $40$ by truncating or adding \texttt{<pad>} tokens. The built vocabulary has $436$ unique tokens thus the vocab size for the word embedding is set to be $436$. For the embedding layers, the feature size is set to $128$ and the feature size of the multi-layer perceptron (MLP) of the encoders and decoders is increased 4 times. In the self-attention layers, 4 heads are used and the encoder and decoder are repeated 6 times. A categorical cross-entropy loss is employed for both the decoder output and the classifier which are added together to compute the final loss. To regularize the network, Dropout with a rate of $0.1$ and layer normalization are employed. The loss function is optimized using the Adam \cite{kingma2014adam} optimizer and trained for a total of 20 epochs. The initial learning rate is set to $10^{-3}$ with an exponential decay learning rate scheduler of the decay rate of $\gamma = 0.95$.



\section{Results} \label{sec:eval}
The proposed network predicts a sequence of most likely pathologies, i.e., DDx, and the actual pathology among the DDx. Both the predicted DDx sequence and the predicted pathology are compared with the ground truth DDx sequence and pathology. The ground truth DDx sequence is organized in descending order of probability distribution of most likely pathologies. As a result, the positional embedding plays an important role in maintaining the correct prediction order leading to a better performance. The ground truth DDx sequence is compared with the predicted sequence elementwise and the mean result is computed. For evaluation, Accuracy, Precision, Recall, and F1 scores are considered. In the following subsections, a comparison of the proposed method with the RL agent-based automated diagnosis methods is performed. Subsequently, the performance of DDx pathology sequence generation and pathology classification will be analyzed and discussed.

\subsection{Comparison}
The baseline models that perform automatic diagnosis using the DDXPlus dataset are Reinforcement Learning (RL)-based agents. Adaptive Alignment of Reinforcement Learning and Classification (AARLC) presented by \cite{yuan2021efficient} is such a system that employs an RL-based agent to adaptively acquire the patient’s symptoms and subsequently uses a classifier to predict the pathology. The process continues iteratively thus generating a DDx sequence of pathologies. Similarly, the baseline automatic symptom detector (BASD) \cite{tchango2022ddxplus} utilizes an RL-based agent to gather evidence and an MLP classifier to predict the pathology. Table \ref{tab:compare} shows the comparison of the performance of the proposed DDxT model with the baseline RL agent-based models.

\begin{table*}[!htbp]
\centering
\caption{Our DDxT improves Precision and thus F1 score significantly, showing value in a generative approach to retrieving accurate diagnoses over the prior RL agent-based approaches. The best results for each metric are highlighted in \textbf{bold}.}
\label{tab:compare}
\vspace{5pt}
\renewcommand{\arraystretch}{1.2}
\adjustbox{max width=0.55\textwidth}{
\begin{tabular}{@{}cccccc@{}}
\toprule
\textbf{Method} & \textbf{GTPA@1} & \textbf{DDP} & \textbf{DDR} & \textbf{DDF1} & \textbf{GM}\\ \midrule
AARLC & 99.21 & 69.53 & \textbf{97.73} & 0.7824 & 87.68 \\ \midrule
BASD  & 97.15 & 88.34 & 85.03 & 0.8369 & 90.03 \\ \midrule
DDxT  & \textbf{99.98} & \textbf{94.84} & 94.65 & \textbf{0.9472} & \textbf{96.45} \\ \bottomrule
\end{tabular}%
}
\end{table*}

The comparison is performed in terms of top-1 ground truth pathology accuracy (GTPA@1), Precision, Recall, and F1 score of DDx denoted as DDP, DDR, and DDF1 following the convention of \cite{tchango2022ddxplus}. AARLC gets the highest recall score but a much lower precision score, lesser than the BASD model, therefore, a lower F1 score. On the other hand, DDxT has a balanced performance in terms of both precision and recall. As a result, it achieved the new highest F1 score of $0.9472$ in the DDXPlus dataset. Additionally, the proposed method outperforms the previous approaches in terms of top-1 accuracy. Moreover, the accuracy, precision, and recall are combined by geometric mean (GM) also shown in Table \ref{tab:compare} to compare the effectiveness of each method. DDxT achieves the best result of $96.45$ with a big margin over the previous RL agent-based methods.

\subsection{DDx Pathology Sequence Generation}
The predicted DDx sequence is compared with the ground truth DDx sequence. Since the ground truth sequence is organized in descending order by the probability distribution, predictions are compared element-wise and the mean result is computed per sequence. To evaluate all the metrics, a confusion matrix is built. In DDx pathology sequence generation, the proposed method achieved $99.82\%$ mean accuracy and a mean F1 score of $ 0.9472\%$. The confusion matrix of the generated pathology sequence is presented in \autoref{ddx_full} which demonstrates the robustness of the proposed generative method. The accuracy, precision, recall, and F1 score of all the pathology classes are also presented in \autoref{ddx_full}. Among all the pathologies, the highest F1 score of $0.9946$ is achieved for \textbf{Myasthenia gravis}, and the minimum F1 score of $0.8643$ is achieved for \textbf{Pancreatic neoplasm}.

\subsection{Pathology Classification}
The proposed network takes the processed feature of the encoder and decoder using a GAP, concatenating them together and feeding them into a classifier for the final pathology classification, i.e., given the list of evidence and set of pathologies (DDx), the final classifier will predict the actual pathology among the most likely DDx pathologies. The results of pathology classification are also evaluated in terms of accuracy, precision, recall, and F1 score. Since the classifier has both encoder and decoder information, it shows significant robustness in classification where the network achieved a mean accuracy of $99.98\%$ with a mean F1 score of $0.9949$. Additionally, the mean precision and recall scores achieved are $99.61\%$ and $99.44\%$, respectively. The minimum F1 score achieved $0.8567$ is for the \textbf{Acute rhinosinusitis}. The confusion matrix of classification along with metric scores for all the pathologies are presented in \autoref{pathology_full}.
\par 
Some conditions, like \textbf{Unstable angina}, \textbf{Acute rhinosinusitis}, and \textbf{Chronic rhinosinusitis} obtain lower precision for varying recall rates. These conditions may need to be considered distinctly in the case of the condition's likelihood to a given population, the risk of the condition itself, and other factors to decide if such conditions are useful to detect in this fashion. Separately, the vast majority of conditions can be detected and a conservative threshold may be used to increase confidence in deployment while expecting a limited reduction in missed diagnoses.



\section{Conclusion and Limitations} \label{sec:conclusion}
In this paper, an automated system of autoregressively generating DDx pathologies and predicting the actual pathology among them is presented. The proposed network uses a Transformer architecture where patient information and evidence are processed by the encoder. Next, the decoder generates a set of likely pathologies. The pathology sequences are generated in the most likely to the least likely order. Afterward, the features of both the encoder and decoder are concatenated and fed to a smaller classifier for the final prediction of the most likely pathology. Experimental results on the DDXPlux dataset demonstrate the feasibility and robustness of DDxT where it achieved a mean accuracy of $99.98\%$ and a mean F1 score of $0.9472$ in the DDx.  Moreover, while predicting the pathology it achieved a mean accuracy of $99.98\%$ with a mean F1 score of $0.9949$. Nevertheless, the proposed system does not acquire and assumes the preexistence of the evidence. Therefore, the performance of the system has a dependency on the correct rendering of accurate information by an authorized user or another automated system. Besides, pathologies such as \textbf{Acute rhinosinusitis}, \textbf{Chronic rhinosinusitis} that displayed lower precision scores with varying recall need to be addressed distinctly. However, in general, the proposed system performed the desired goal of automating differential diagnosis and has the potential to emerge as an assistive tool for the physician during the diagnosis process.

\bibliographystyle{plain}

\appendix
\newpage

\section{Transformer} \label{a:appendix}
Transformer \cite{vaswani2017attention} is an attention-based model that uses its input to generate query $\mathbf{Q}$, key $\mathbf{K}$, and value $\mathbf{V}$ to compute self-attention given in Equation \ref{attention} where $d$ is the feature dimension. The self-attention layer is accompanied by multi-layer perceptron, layer normalization, and skip connections to form an attention block. The blocks are repeated multiple times. The sequence of the patient’s information will be processed on a stack of blocks called the encoder which gives context to the decoder, another stack of decoder blocks that produces the output. A look-ahead mask is applied to the decoder to generate output autoregressively so that the current prediction $p_{t}$ only relies on the previous predictions from $p_{0}$ to $p_{t-1}$.
\begin{equation}
\texttt{Attention(}\mathbf{Q}, \mathbf{K}, \mathbf{V} \texttt{)} = \texttt{softmax}\left(\frac{\mathbf{Q}\mathbf{K}^{T}}{\sqrt{d}}\right) \mathbf{V}
\label{attention}
\end{equation}

\section{DDx Comprehensive Results} \label{ddx_full}
\begin{figure}[!htbp]
\centerline{\includegraphics[width=\columnwidth]{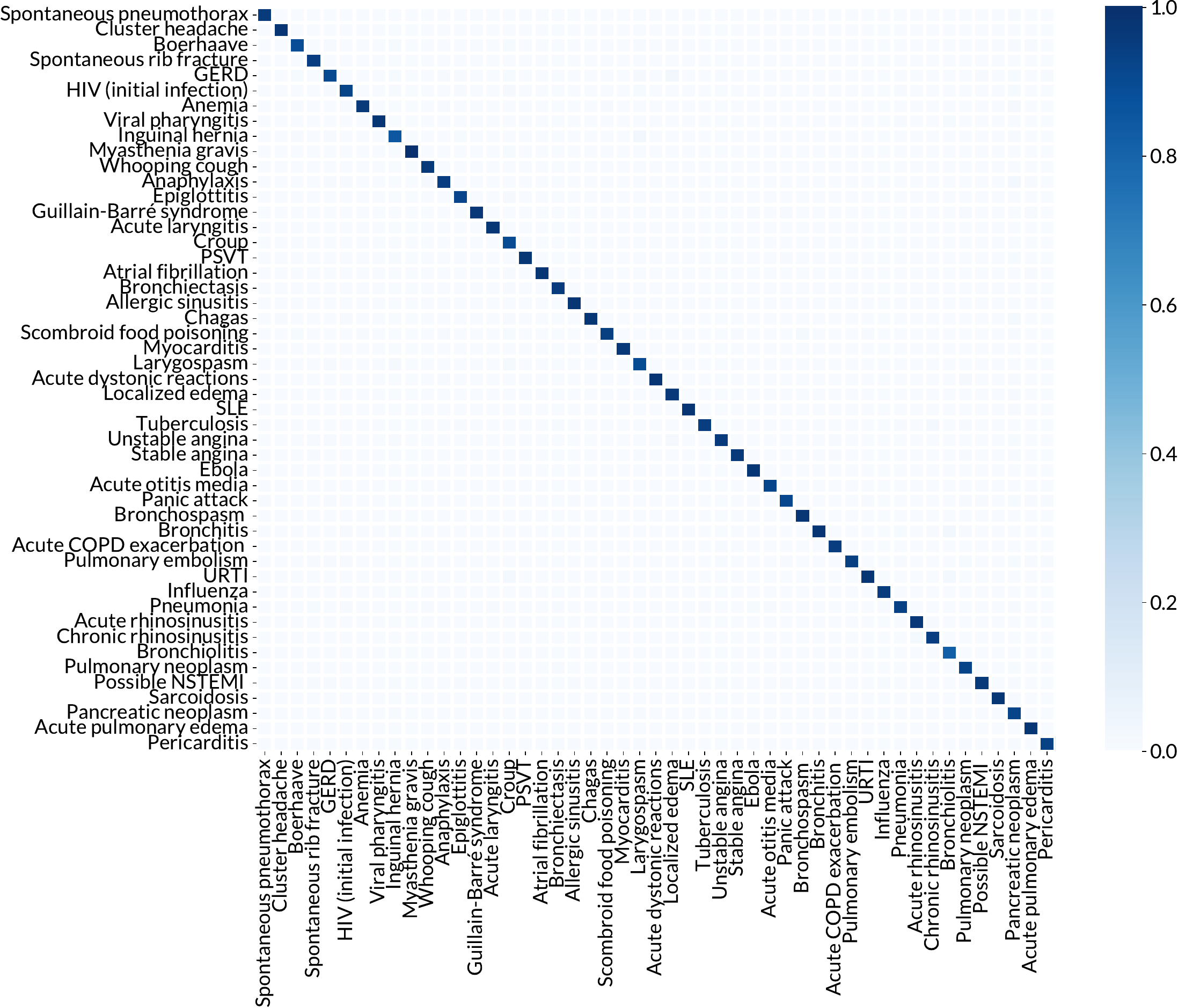}}
\caption{Confusion matrix for the DDx sequence generation where each element, i.e, pathology of the sequence is evaluated elementwise. Each row indicates the ground truth class and each column indicates the predicted class.} 
\label{fig:conf_mat_ddx}
\end{figure}

The confusion matrix for the DDx sequence generation is presented in \autoref{fig:conf_mat_ddx}. The dataset has a total of $49$ pathologies. So, the confusion matrix is of the size of $49 \times 49$. The ground truth classes are shown in the row and the predicted labels are in the column. Visually, the proposed method has very few false positives and false negatives while generating the most likely pathologies.

The accuracy, precision, recall, and F1 score of all the pathology classes are also presented in \autoref{tab:ddx_results}. Among all the pathologies, the highest F1 score of $0.9946$ is achieved for \textbf{Myasthenia gravis}, and the lowest F1 score of $0.8643$ is achieved for \textbf{Pancreatic neoplasm}.

\begin{table}[!htbp]
\centering
\caption{Classification results of the Differential Diagnosis (DDx) where the sequence of pathologies are generated given a set of evidence.}
\label{tab:ddx_results}
\renewcommand{\arraystretch}{0.95}
\adjustbox{max width=\textwidth}{
\begin{tabular}{|c|c|c|c|c|}
\hline
\textbf{Pathology} & \textbf{Acc. (\%)} & \textbf{Prec. (\%)} & \textbf{Rec. (\%)} & \textbf{F1} \\ \hline
Spontaneous pneumothorax & 99.91 & 96.11 & 95.65 & 0.9588 \\ \hline
Cluster headache & 99.92 & 96.69 & 97.73 & 0.9720 \\ \hline
Boerhaave & 99.77 & 93.04 & 89.01 & 0.9098 \\ \hline
Spontaneous rib fracture & 99.95 & 93.07 & 94.38 & 0.9372 \\ \hline
GERD & 99.56 & 91.54 & 90.44 & 0.9098 \\ \hline
HIV (initial infection) & 99.67 & 91.28 & 92.30 & 0.9179 \\ \hline
Anemia & 99.67 & 96.89 & 96.46 & 0.9668 \\ \hline
Viral pharyngitis & 99.94 & 98.44 & 97.57 & 0.9800 \\ \hline
Inguinal hernia & 99.84 & 90.58 & 84.94 & 0.8767 \\ \hline
Myasthenia gravis & 99.96 & 99.53 & 99.40 & 0.9946 \\ \hline
Whooping cough & 99.99 & 97.67 & 95.87 & 0.9676 \\ \hline
Anaphylaxis & 99.59 & 94.68 & 94.42 & 0.9455 \\ \hline
Epiglottitis & 99.92 & 94.13 & 91.81 & 0.9296 \\ \hline
Guillain-Barré syndrome & 99.78 & 97.43 & 96.83 & 0.9713 \\ \hline
Acute laryngitis & 99.92 & 96.93 & 97.10 & 0.9701 \\ \hline
Croup & 99.93 & 85.13 & 89.49 & 0.8726 \\ \hline
PSVT & 99.86 & 98.38 & 97.20 & 0.9778 \\ \hline
Atrial fibrillation & 99.85 & 98.26 & 98.03 & 0.9815 \\ \hline
Bronchiectasis & 99.85 & 94.42 & 95.06 & 0.9474 \\ \hline
Allergic sinusitis & 99.99 & 96.73 & 97.47 & 0.9710 \\ \hline
Chagas & 99.69 & 97.05 & 96.91 & 0.9698 \\ \hline
Scombroid food poisoning & 99.56 & 95.13 & 93.94 & 0.9453 \\ \hline
Myocarditis & 99.78 & 97.01 & 96.76 & 0.9689 \\ \hline
Larygospasm & 99.92 & 91.21 & 89.60 & 0.9040 \\ \hline
Acute dystonic reactions & 99.86 & 98.81 & 97.78 & 0.9829 \\ \hline
Localized edema & 99.95 & 94.54 & 95.76 & 0.9515 \\ \hline
SLE & 99.92 & 98.28 & 98.01 & 0.9814 \\ \hline
Tuberculosis & 99.76 & 95.39 & 94.62 & 0.9500 \\ \hline
Unstable angina & 99.63 & 93.21 & 95.54 & 0.9436 \\ \hline
Stable angina & 99.76 & 95.28 & 96.31 & 0.9579 \\ \hline
Ebola & 99.98 & 94.62 & 96.96 & 0.9578 \\ \hline
Acute otitis media & 99.95 & 96.46 & 91.67 & 0.9400 \\ \hline
Panic attack & 99.65 & 94.24 & 91.40 & 0.9280 \\ \hline
Bronchospasm  & 99.92 & 95.42 & 97.52 & 0.9646 \\ \hline
Bronchitis & 99.85 & 98.48 & 97.70 & 0.9809 \\ \hline
Acute COPD exacerbation  & 99.95 & 96.88 & 94.66 & 0.9576 \\ \hline
Pulmonary embolism & 99.68 & 96.89 & 94.13 & 0.9549 \\ \hline
URTI & 99.88 & 96.28 & 97.69 & 0.9698 \\ \hline
Influenza & 99.81 & 94.05 & 95.11 & 0.9458 \\ \hline
Pneumonia & 99.67 & 92.32 & 93.32 & 0.9282 \\ \hline
Acute rhinosinusitis & 99.95 & 96.64 & 96.69 & 0.9667 \\ \hline
Chronic rhinosinusitis & 99.90 & 94.25 & 94.52 & 0.9438 \\ \hline
Bronchiolitis & 100.00 & 92.59 & 81.97 & 0.8696 \\ \hline
Pulmonary neoplasm & 99.73 & 89.91 & 92.44 & 0.9116 \\ \hline
Possible NSTEMI  & 99.60 & 94.99 & 95.73 & 0.9536 \\ \hline
Sarcoidosis & 99.90 & 97.90 & 96.90 & 0.9740 \\ \hline
Pancreatic neoplasm & 99.56 & 81.10 & 92.52 & 0.8643 \\ \hline
Acute pulmonary edema & 99.80 & 94.53 & 97.47 & 0.9598 \\ \hline
Pericarditis & 99.72 & 92.57 & 93.25 & 0.9291 \\ \hline \hline
\textbf{Mean} & 99.82 & 94.84 & 94.65 & 0.9472 \\ \hline
\end{tabular}
}
\end{table}

\section{Pathology Classification Comprehensive Results} \label{pathology_full}
\begin{figure}[!htbp]
\centerline{\includegraphics[width=\columnwidth]{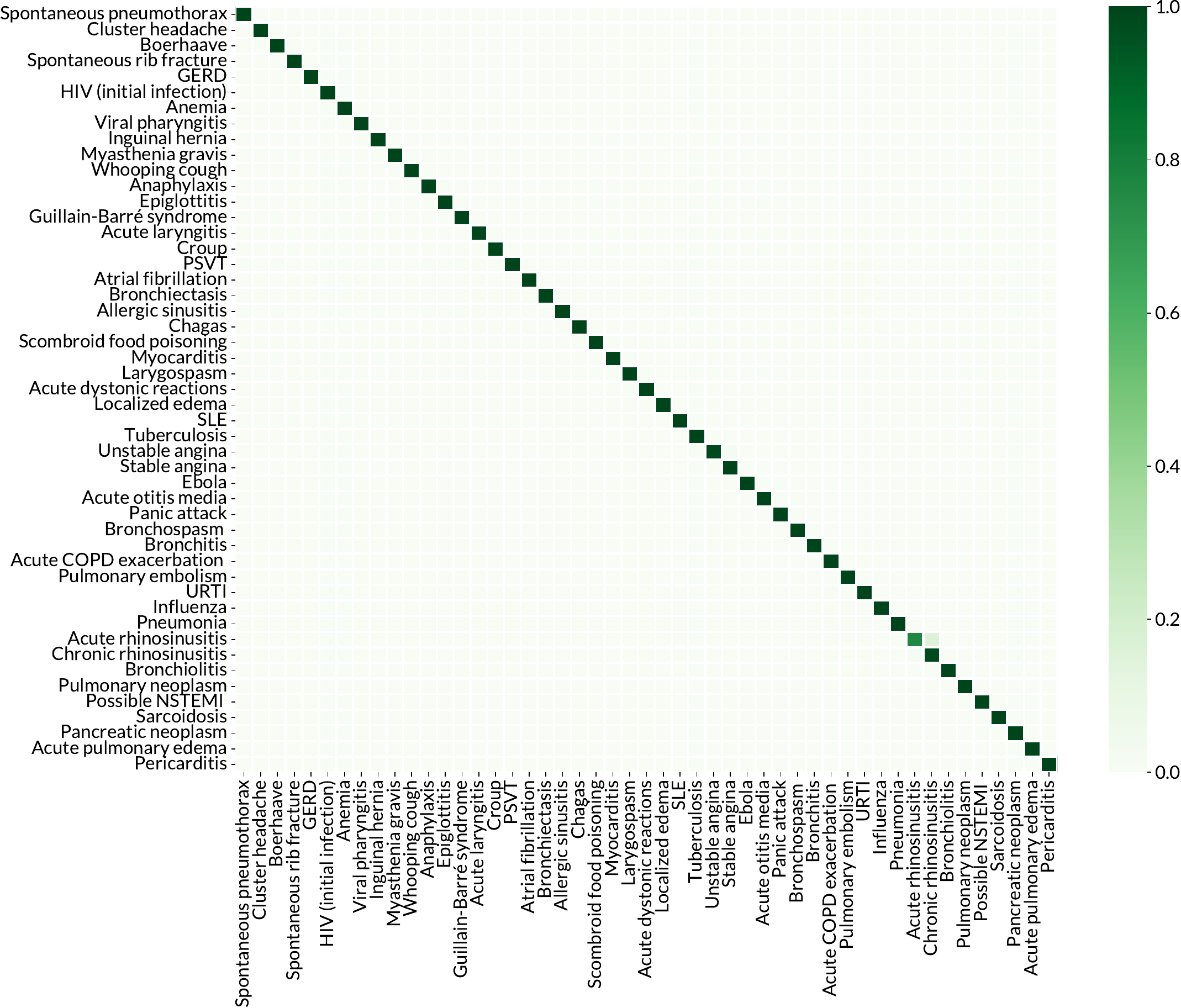}}
\caption{Confusion matrix of the final pathology classification. True labels are shown in the row and the predicted labels are shown in the column.} 
\label{fig:conf_mat_cls}
\end{figure}

The confusion matrix of the pathology classification is presented in \autoref{fig:conf_mat_cls}. The proposed method showed a robust performance while classifying pathology from encoder and decoder features which can be observed from the figure.

\newpage

The proposed method has achieved a mean accuracy of $99.98\%$ with a mean F1 score of $0.9949$. Additionally, the mean precision and recall scores achieved are $99.61\%$ and $99.44\%$, respectively. The accuracy, precision, recall, and F1 score of all the classes for pathology classification are also presented in \autoref{tab:cls_result}.

\begin{table}[!htbp]
\centering
\caption{Results of the pathology classification the classifier processed the encoder and the decoder information to predict the most likely pathology among the predicted set of differential diagnoses.}
\label{tab:cls_result}
\renewcommand{\arraystretch}{0.95}
\adjustbox{max width=\textwidth}{
\begin{tabular}{|c|c|c|c|c|}
\hline
\textbf{Pathology} & \textbf{Acc. (\%)} & \textbf{Prec. (\%)} & \textbf{Rec. (\%)} & \textbf{F1} \\ \hline
Spontaneous pneumothorax & 100.00 & 100.00 & 100.00 & 1.0000 \\ \hline
Cluster headache & 100.00 & 100.00 & 100.00 & 1.0000 \\ \hline
Boerhaave & 100.00 & 100.00 & 100.00 & 1.0000 \\ \hline
Spontaneous rib fracture & 100.00 & 100.00 & 100.00 & 1.0000 \\ \hline
GERD & 100.00 & 100.00 & 100.00 & 1.0000 \\ \hline
HIV (initial infection) & 100.00 & 100.00 & 100.00 & 1.0000 \\ \hline
Anemia & 100.00 & 100.00 & 100.00 & 1.0000 \\ \hline
Viral pharyngitis & 99.97 & 99.72 & 99.77 & 0.9975 \\ \hline
Inguinal hernia & 100.00 & 100.00 & 100.00 & 1.0000 \\ \hline
Myasthenia gravis & 100.00 & 100.00 & 100.00 & 1.0000 \\ \hline
Whooping cough & 100.00 & 100.00 & 100.00 & 1.0000 \\ \hline
Anaphylaxis & 100.00 & 100.00 & 100.00 & 1.0000 \\ \hline
Epiglottitis & 100.00 & 100.00 & 100.00 & 1.0000 \\ \hline
Guillain-Barré syndrome & 100.00 & 100.00 & 100.00 & 1.0000 \\ \hline
Acute laryngitis & 99.97 & 99.41 & 99.29 & 0.9935 \\ \hline
Croup & 100.00 & 100.00 & 100.00 & 1.0000 \\ \hline
PSVT & 100.00 & 100.00 & 100.00 & 1.0000 \\ \hline
Atrial fibrillation & 100.00 & 100.00 & 100.00 & 1.0000 \\ \hline
Bronchiectasis & 100.00 & 100.00 & 100.00 & 1.0000 \\ \hline
Allergic sinusitis & 100.00 & 100.00 & 100.00 & 1.0000 \\ \hline
Chagas & 100.00 & 100.00 & 100.00 & 1.0000 \\ \hline
Scombroid food poisoning & 100.00 & 100.00 & 100.00 & 1.0000 \\ \hline
Myocarditis & 100.00 & 100.00 & 100.00 & 1.0000 \\ \hline
Larygospasm & 100.00 & 100.00 & 100.00 & 1.0000 \\ \hline
Acute dystonic reactions & 100.00 & 100.00 & 100.00 & 1.0000 \\ \hline
Localized edema & 100.00 & 100.00 & 100.00 & 1.0000 \\ \hline
SLE & 100.00 & 100.00 & 100.00 & 1.0000 \\ \hline
Tuberculosis & 100.00 & 100.00 & 100.00 & 1.0000 \\ \hline
Unstable angina & 99.96 & 99.96 & 98.37 & 0.9916 \\ \hline
Stable angina & 99.97 & 98.07 & 100.00 & 0.9902 \\ \hline
Ebola & 100.00 & 100.00 & 100.00 & 1.0000 \\ \hline
Acute otitis media & 100.00 & 100.00 & 100.00 & 1.0000 \\ \hline
Panic attack & 100.00 & 100.00 & 100.00 & 1.0000 \\ \hline
Bronchospasm  & 100.00 & 100.00 & 100.00 & 1.0000 \\ \hline
Bronchitis & 100.00 & 100.00 & 100.00 & 1.0000 \\ \hline
Acute COPD exacerbation  & 100.00 & 100.00 & 100.00 & 1.0000 \\ \hline
Pulmonary embolism & 100.00 & 100.00 & 100.00 & 1.0000 \\ \hline
URTI & 100.00 & 100.00 & 100.00 & 1.0000 \\ \hline
Influenza & 100.00 & 100.00 & 100.00 & 1.0000 \\ \hline
Pneumonia & 100.00 & 100.00 & 100.00 & 1.0000 \\ \hline
Acute rhinosinusitis & 99.65 & 97.36 & 76.49 & 0.8567 \\ \hline
Chronic rhinosinusitis & 99.65 & 86.30 & 98.62 & 0.9205 \\ \hline
Bronchiolitis & 100.00 & 100.00 & 100.00 & 1.0000 \\ \hline
Pulmonary neoplasm & 100.00 & 100.00 & 100.00 & 1.0000 \\ \hline
Possible NSTEMI  & 100.00 & 100.00 & 99.97 & 0.9998 \\ \hline
Sarcoidosis & 100.00 & 100.00 & 100.00 & 1.0000 \\ \hline
Pancreatic neoplasm & 100.00 & 100.00 & 100.00 & 1.0000 \\ \hline
Acute pulmonary edema & 100.00 & 100.00 & 100.00 & 1.0000 \\ \hline
Pericarditis & 100.00 & 100.00 & 100.00 & 1.0000 \\ \hline \hline
\textbf{Mean} & 99.98 & 99.61 & 99.44 & 0.9949 \\ \hline
\end{tabular}
}
\end{table}

\end{document}